\documentclass[letterpaper, 10 pt, journal, twoside]{ieeetran}

\usepackage{graphics}
\usepackage{amsmath}
\usepackage{amssymb}
\usepackage{amsfonts}
\usepackage{algorithm}
\usepackage{algorithmic}
\usepackage{array}
\usepackage{booktabs}
\usepackage{colortbl}
\usepackage{footmisc}
\usepackage{graphicx}
\usepackage{makecell}
\usepackage{mathrsfs}
\usepackage{mathtools}
\usepackage{multicol}
\usepackage{multirow}
\usepackage{nicematrix}
\usepackage{subcaption}
\usepackage{xcolor}

\definecolor{Gray}{gray}{0.95}

\newcommand{\vect}[1]{\boldsymbol{\mathbf{#1}}}
\newcolumntype{L}[1]{>{\raggedright\arraybackslash}p{#1}}
\newcolumntype{C}[1]{>{\centering\arraybackslash}p{#1}}

\begin{document}

\title{Efficient and Safe Trajectory Planning for Autonomous Agricultural Vehicle Headland Turning in Cluttered Orchard Environments}

\author{Peng Wei$^{2}$, Chen Peng$^{1,3}$, Wenwu Lu$^{1}$, Yuankai Zhu$^{4}$, Stavros Vougioukas$^{2}$, Zhenghao Fei$^{1,3}$, and Zhikang Ge$^{1}$%
\thanks{Accepted version. © 2025 IEEE.  Personal use of this material is permitted.  Permission from IEEE must be obtained for all other uses, in any current or future media, including reprinting/republishing this material for advertising or promotional purposes, creating new collective works, for resale or redistribution to servers or lists, or reuse of any copyrighted component of this work in other works. This work was supported by the National Science Foundation of Zhejiang Province, China (Grant NO. LD24C130003) \textit{(Corresponding author: Chen Peng)}.}
\thanks{$^{1}$ Chen Peng, Wenwu Lu, Zhenghao Fei, and Zhikang Ge are with the ZJU-Hangzhou Global Scientific and Technological Innovation Center, Zhejiang University, Hangzhou, China
        {\tt\footnotesize \{chen.peng, wenwuLu, zfei, zge\}@zju.edu.cn}}%
\thanks{$^{2}$ Peng Wei and Stavros Vougioukas are with the Department of Biological and Agricultural Engineering, University of California, Davis,
        One Shields Ave, Davis, USA
        {\tt\footnotesize \{penwei, svougioukas\}@ucdavis.edu}}%
\thanks{$^{3}$ Chen Peng and Zhenghao Fei are with the College Of Biosystems Engineering And Food Science, Zhejiang University,
        Hangzhou, China
        {\tt\footnotesize \{chen.peng, zfei\}@zju.edu.cn}}%
\thanks{$^{4}$ Yuankai Zhu is with the Department of Mechanical and Aerospace Engineering, University of California, Davis,
        One Shields Ave, Davis, USA
        {\tt\footnotesize ykzhu@ucdavis.edu}}%
}

\markboth{THIS PAPER HAS BEEN ACCEPTED FOR PUBLICATION AT IEEE ROBOTICS AND AUTOMATION LETTERS (RA-L). © IEEE}
{Wei \MakeLowercase{\textit{et al.}}: Efficient and Safe Trajectory Planning for Autonomous Agricultural Vehicle Headland Turning} 

\maketitle

\begin{abstract}
Autonomous agricultural vehicles (AAVs), including field robots and autonomous tractors, are becoming essential in modern farming by improving efficiency and reducing labor costs. A critical task in AAV operations is headland turning between crop rows. This task is challenging in orchards with limited headland space, irregular boundaries, operational constraints, and static obstacles. While traditional trajectory planning methods work well in arable farming, they often fail in cluttered orchard environments. This letter presents a novel trajectory planner that enhances the safety and efficiency of AAV headland maneuvers, leveraging advancements in autonomous driving. Our approach includes an efficient front-end algorithm and a high-performance back-end optimization. Applied to vehicles with various implements, it outperforms state-of-the-art methods in both standard and challenging orchard fields. This work bridges agricultural and autonomous driving technologies, facilitating a broader adoption of AAVs in complex orchards.
\end{abstract}


\section{Introduction}
\label{sec:introduction}
\IEEEPARstart{I}{n} modern farming, autonomous agricultural vehicles (AAVs) have gained significant popularity~\cite{Fasiolo2023}. These vehicles navigate through crop rows and perform headland maneuvers to transition between rows, allowing them to cover the entire field~\cite{Mier2023}. Effective trajectory planning is crucial to ensure safe and efficient farm operations. In arable farming, the absence of dense obstacles simplifies trajectory planning, while the flexible cultivation layout eases the replanting of crops. These factors considerably facilitate automation and reduce operational challenges for AAVs.

However, trajectory planning presents difficulties in many orchards, particularly during headland turning. Local topography and the desire to maximize land use can result in narrow headlands with irregular shapes. Stationary obstacles (e.g., irrigation pipes, utility poles) and the complex geometries of vehicles with their implements~\cite{Peng2024} present additional challenges. Thus, headland turning can be complex, requiring a balance of safety and mission efficiency under the operational constraints of non-holonomic vehicles. An AAV will perform hundreds of headland turns, and suboptimal trajectories can introduce delays and increase costs, as turning maneuvers are non-productive and consume additional time and energy~\cite{Vougioukas2019}. Additionally, a dynamically infeasible trajectory cannot be accurately tracked by the vehicle and may lead to collisions. Furthermore, agricultural vehicles typically operate on rough terrain, which increases the likelihood of deviations from planned trajectories and necessitates frequent replanning. Therefore, a planner that can efficiently generate viable and optimal trajectories in constrained headlands is highly desirable.

\begin{figure}[t]
\centering
    \includegraphics[width=0.85\linewidth]{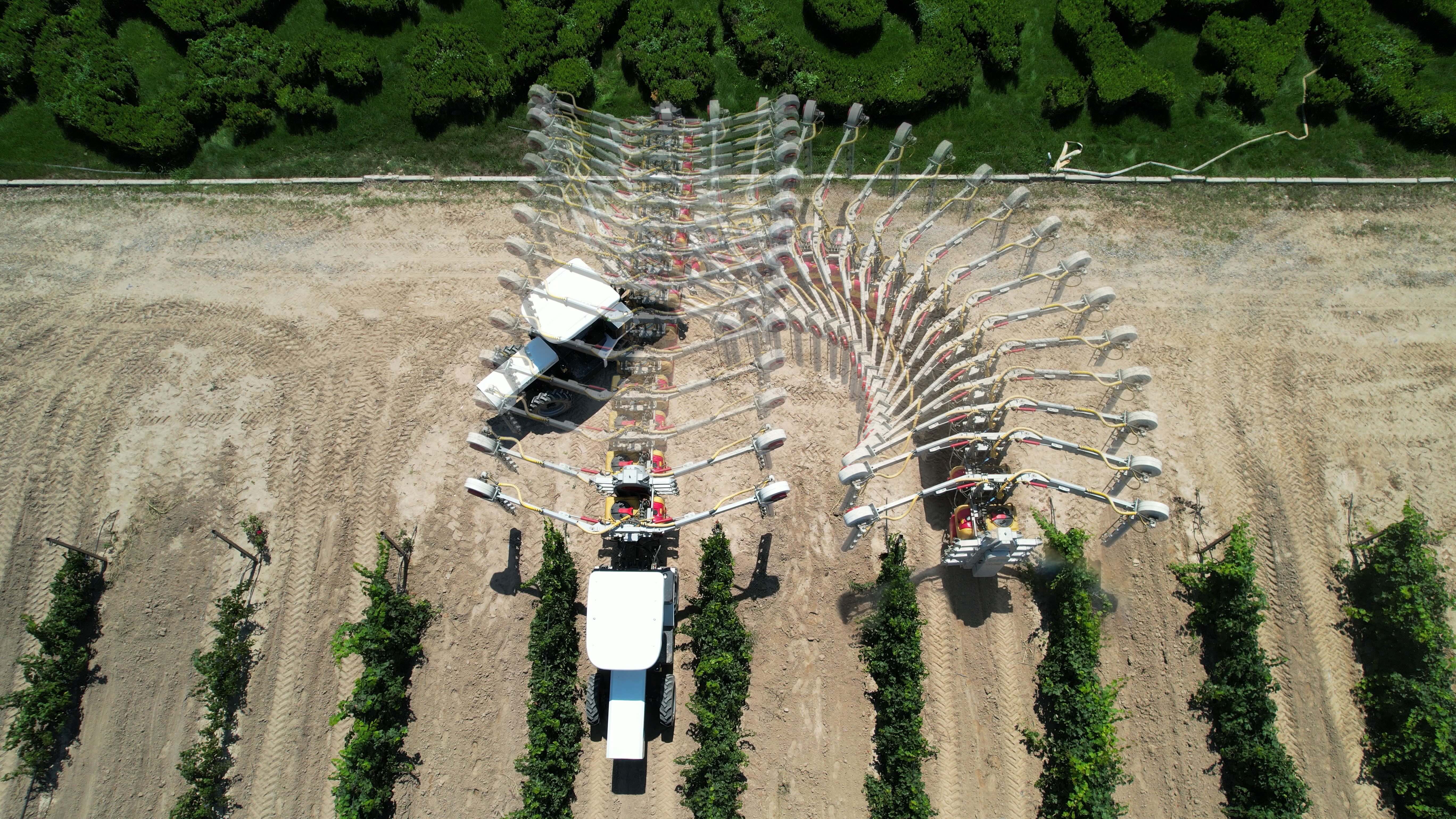}
    \caption{Illustration of an electric tractor equipped with a pruner executing a turning trajectory in a headland.}
\label{fig:field_tractor}
\end{figure}

This work focuses on AAV trajectory planning in headland turning scenarios, as illustrated in Fig.~\ref{fig:field_tractor}. We propose a practical trajectory planner to ensure safe and optimal maneuvers in cluttered headland spaces. The planner follows a two-stage framework. In the front end, we introduce an enhanced hybrid A* method that accelerates collision checking, leading to faster initial trajectory generation tailored for headland turning. In the back end, a high-efficiency module refines the trajectory to satisfy kinematic and collision constraints while optimizing the control effort and trajectory duration. By employing multiple safe corridors, our approach reduces the conservatism caused by the complex geometries of agricultural vehicles with various implements. Experimental results show that our planner significantly improves computational efficiency and success rates compared to state-of-the-art methods. The main contributions of our work are summarized as follows:
\begin{enumerate}
    \item We propose an optimization-based trajectory planner for AAV headland turning that effectively accounts for the geometries of attached implements and the operational constraints in cluttered environments.
    \item We develop an advanced collision detection method tailored to headland turning scenarios, significantly improving the efficiency of front-end trajectory generation while maintaining accuracy.
    \item We introduce an efficient back-end optimization algorithm that refines front-end trajectories to satisfy safety and vehicle constraints. By utilizing multiple safe corridors, our approach mitigates conservatism from complex vehicle geometries, achieving a balance between accuracy and efficiency.
    \item Our approach outperforms state-of-the-art methods for AAV headland turning in computation time and success rates and generates more optimal trajectories in tight headlands with complex vehicle geometries.
\end{enumerate}

\section{Related Work}
\label{sec:related_work}

\subsection{Headland Turning for Agricultural Vehicles}
\label{sec:related_work_headland_turning}
Agricultural vehicles perform numerous headland turns, contributing significantly to non-productive time. Researchers have focused on optimizing turning paths or trajectories, considering both agronomic and economic aspects. Traditional methods, such as Dubins and Reeds-Shepp curves~\cite{Dubins1957, Reeds1990}, generate turning paths without considering actuator dynamics, leading to abrupt steering and velocity changes that hinder accurate path tracking. Backman et al. ~\cite{Backman2015} incorporated actuator dynamics into trajectory generation through a numerical integration approach to address this issue. Similarly, Sabelhaus et al.\cite{Sabelhaus2013} proposed a continuous-curvature path planning method using clothoids to produce smoother transitions. Additionally, He et al.~\cite{He2023} developed a dynamic path planning framework to handle frequent replanning due to vehicle side slips. Although effective in regular orchards, these methods are prone to failure in constrained headlands and can produce unsafe trajectories in fields with irregular boundaries and narrow spaces~\cite{Peng2024}.

In response, many researchers have treated headland trajectory planning as an optimal control problem (OCP), optimizing a cost function while meeting kinematic and obstacle-avoidance constraints. For example, Oksanen~\cite{Oksanen2004} explored minimum-time tractor-trailer planning in headlands, and Tu and Tang~\cite{Tu2019} proposed a direct optimization approach considering tractor-implement kinematics. Vougioukas et al.~\cite{Vougioukas2006} introduced a two-stage planner for headland turns, considering static obstacles. Although effective in simpler orchards, these methods may fail in more complex environments and do not demonstrate robustness when applied to vehicles with various implements.

\subsection{Autonomous Vehicle Parking}
\label{sec:related_work_autonomous_parking}

In the literature on autonomous driving, parking trajectory planning is analogous to headland turning in agriculture, which has been extensively studied. Graph search-based methods, such as hybrid A*~\cite{Dolgov2010}, and sampling-based methods, like the kinodynamic Rapidly-exploring Random Tree (RRT)~\cite{Webb2013}, are commonly used to generate collision-free paths considering vehicle dynamics. Although simple and effective, these methods often require additional steps to calculate speed profiles, leading to infeasible trajectories. While there are other techniques~\cite{Paden2016}, this work focuses on optimization-based approaches given their advantages in complex environments.

Optimization-based methods are favored for their ability to converge to locally optimal solutions with theoretical guarantees. Kondak and Hommel~\cite{Kondak2001} first formulated parking trajectory generation as an OCP and successfully solved it using a numerical approach. Li et al.~\cite{Li2016} addressed time-optimal maneuver planning for parallel parking through dynamic optimization techniques, while Li et al.~\cite{Li2015} proposed a unified motion planner that extended beyond standard rectangular parking scenarios to handle irregularly placed obstacles. However, both methods suffer from long computation times and were only validated through simulations, which limits their practical application. To enhance efficiency and robustness, some researchers have adopted a hierarchical framework~\cite{Zhou2021}, where a front-end algorithm generates a collision-free trajectory that is subsequently refined in the back end to optimize a cost function while satisfying various constraints. While adequate for autonomous driving, this approach is not suitable for agricultural vehicles with complex shapes. Given that collision avoidance constraints are often non-convex and non-differentiable, Zhang et al.~\cite{Zhang2018} introduced an optimization-based collision avoidance (OBCA) algorithm to improve the differentiability of these constraints. This method is adaptable to complex vehicle geometries and ensures obstacle avoidance and smooth trajectories. However, despite its advantages, the method scales poorly with the number of obstacles due to the computational and memory demands associated with the dual variables.

To address computational efficiency,  corridor-based approaches have been proposed to limit the search space within a local region~\cite{Li2021, Han2023}. These approaches effectively decouple the dimensionality of the collision-free constraints from the number of obstacles, enabling faster local solutions. Furthermore, Han et al.~\cite{Han2023} proposed an efficient planner that jointly optimizes the spatial and temporal properties of trajectories in unstructured environments, demonstrating real-time performance. However, these methods simplify the robot to one or two circles or represent the vehicle as a simple rectangle. Such abstractions, while computationally advantageous, introduce conservatism to trajectory planning. As a result, it can lead to overly conservative trajectories or infeasible solutions when applied to agricultural vehicles with various implements in farm operations.

\section{Safe and Efficient Headland Trajectory Planning}
\label{sec:methodology}

This section presents our trajectory planner for headland turning in constrained environments, using geometric primitives to represent the environment and the AAV. A two-stage optimization approach is employed: the front end uses an enhanced hybrid A* algorithm to generate a collision-free initial trajectory, while the back end optimizes it for smoothness, duration, and constraint satisfaction.

\begin{figure}[t]
\centering
    \includegraphics[width=0.85\linewidth]{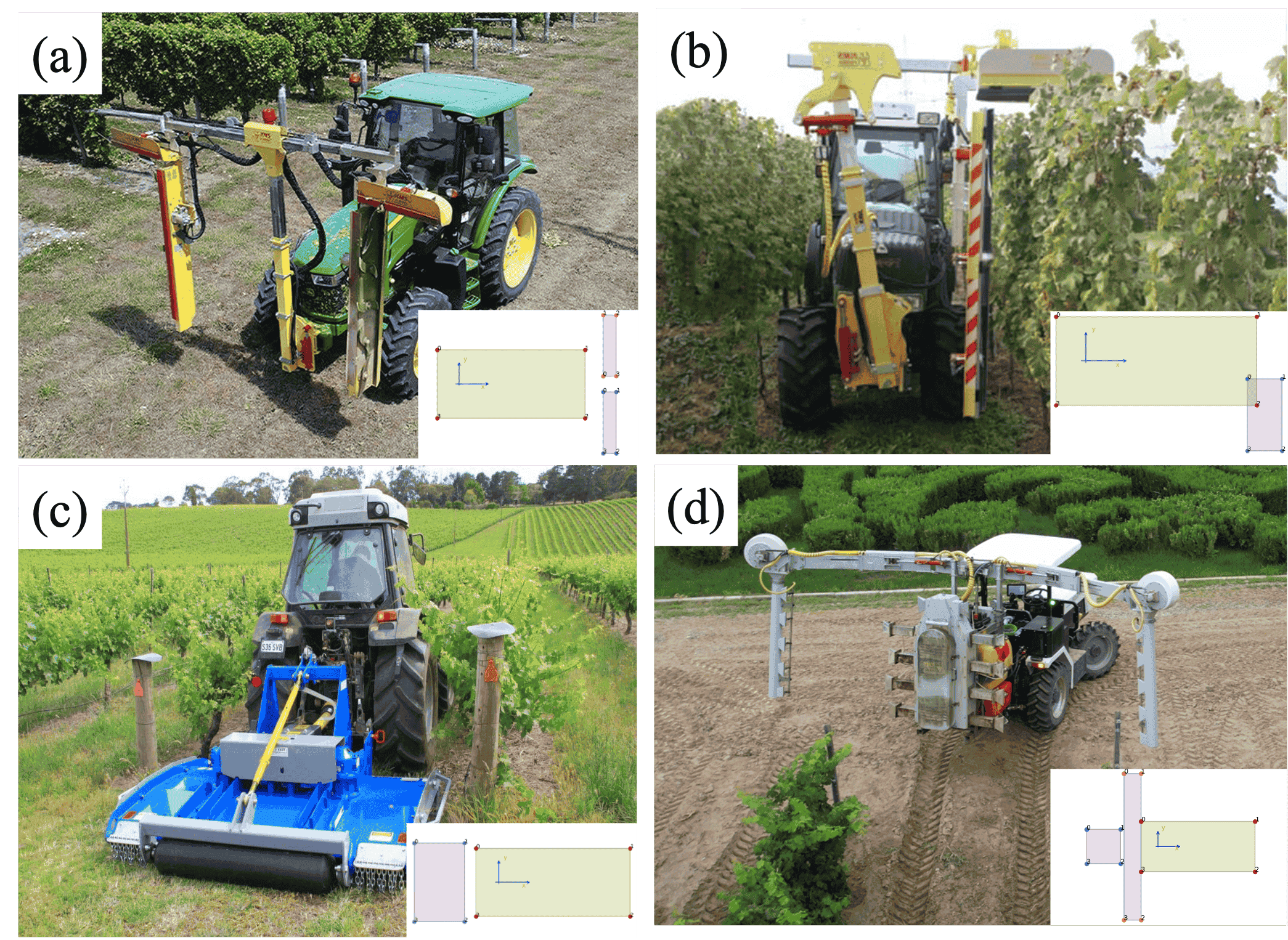}
    \caption{Vineyard tractors with: (a) double-sided pruner, (b) single-sided pruner, (c) mower, and (d) KMS sprayer, along with their geometric representations for trajectory planning.}
\label{fig:tractor_implement}
\end{figure}

\subsection{Geometric Representations}
\label{sec:method_geometric_representation}

Headland obstacles typically include field boundaries, crop rows, and other stationary objects. In this research, we consider these obstacles to be time-invariant in shape and location, and their occupied spaces are non-traversable by the AAV (including the vehicle body and implement). We represent these obstacles using $N_o \in \mathbb{Z}^+$ polygons, denoted $\mathbb{O}_1,\ldots,\mathbb{O}_{N_o} \subset \mathbb{R}^2$. Traditional approaches often model the AAV as a convex hull, which can result in overly conservative representations and cause failure in collision detection, especially when the AAV starts inside a crop row. Instead, we model the vehicle and its implements using $N_v = N_{veh} + N_{imp} \in \mathbb{Z}^+$ \textit{rigidly connected rectangles}, where $N_{veh}$ ($N_{imp}$) represents the number of vehicle (implement) parts. These rectangles, denoted as $\mathbb{V}_1(t),\ldots,\mathbb{V}_{N_v}(t) \subset \mathbb{R}^2$, have positions and orientations that vary over time. These geometric abstractions are widely accepted in many agricultural applications~\cite{Oksanen2009}, and we assume they are provided before trajectory planning. To ensure collision avoidance, the intersection of the occupied spaces of the vehicle-implement rectangles and the obstacle polygons must remain empty at all times,
$\mathbb{O}_i \cap \mathbb{V}_j(t)=\emptyset, \forall i\in \{1,\ldots,N_o\}, \forall j\in \{1,\ldots,N_v\}, \forall t\in \mathbb{R}^+$.

In Fig.~\ref{fig:tractor_implement}, we depict four common vehicle-implement systems considered in this work, operating in vineyards along with their geometric representations\footnote{We modeled the overhanging implements in Fig.~\ref{fig:tractor_implement}(a),(b), and (d) as single long rectangles for simplicity. This approach is valid because we only consider trajectories during headland turns.}.

\subsection{Efficient Collision Detection for Heading Turning}
\label{sec:method_collision_detection}

We employ a hybrid A* algorithm as the front end of our trajectory planner. One bottleneck in hybrid A* search is the running time of collision checking. To improve efficiency, we propose a covering circle inflation method adapted from the hierarchical map inflation technique in \cite{Zheng2023}. Our approach computes covering circles for the vehicle and attached implement, then inflates the obstacles on a grid map. The map is derived from the previous geometric representation and discretized with a resolution of $\delta$. By converting full-body collision detection into finite-point checking in the configuration space, we reduce the complexity of collision checking to linear time, significantly improving runtime performance.

\begin{figure}[t]
    \centering
    \includegraphics[width=0.8\linewidth]{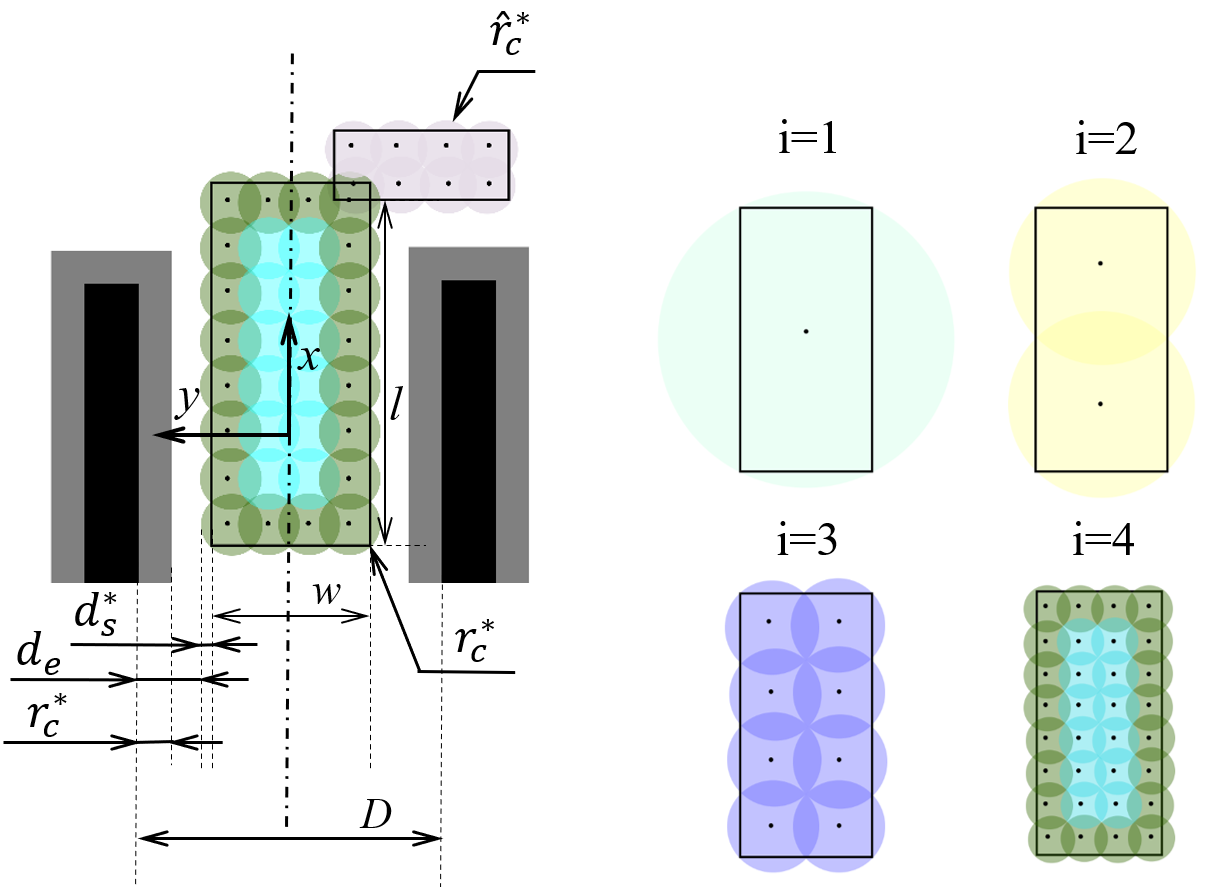}
    \caption{Process of determining optimal covering circles for vehicle and implement. The black areas represent trees, and the gray areas represent map inflation.}
    \label{fig:circle_cover_and_max}
\end{figure}

In headland turning, the most constrained situation occurs when the AAV operates within crop rows. Therefore, we compute the covering circles $\mathscr{C}(\vect{p}_c, r_c)$ under this condition iteratively, as illustrated in Fig.~\ref{fig:circle_cover_and_max}. Starting with the vehicle body and assuming $N_{veh}=1$ (i.e., as a single rectangle), the radius $r_c^i$ of the covering circle at each iteration $i\in \mathbb{Z}^+$ is calculated as: 
\begin{equation}
    r_c^i = \sqrt{(\frac{l}{2^i})^2 + (\frac{w}{2^{max(1,i-1)}}})^2
\label{eq:split_radius}
\end{equation}
where $l$ and $w$ represent the vehicle length and width. The center position of the bottom-left circle in the body frame is computed as:
\begin{equation}
    \vect{p}_{c,1}^i = \big[x_{bl} + \frac{l}{2^i},~y_{bl} + \frac{w}{2^{max(1,i-1)}}\big]^T
\label{eq:center_coordinates_corrected}
\end{equation}
where $[x_{bl},y_{bl}]^T$ are the coordinates of the bottom-left corner of the vehicle's rectangle in the body frame. The centers of the other circles are computed incrementally (for $i>1$) using the offsets $(\Delta x^i, \Delta y^i) = (l/2^{i-1},w/2^{i-2})$, until $2^{i-1}$ and $2^{i-2}$ circles are achieved in the $x$ and $-y$ directions, respectively. The overhang distance $d_s^i$, which measures how much the covering circle extends beyond the vehicle's rectangle, can be calculated for each circle as: 
\begin{equation}
    d_s^i = r_c^i - \min(\frac{l}{2^i},~\frac{w}{2^{max(1,i-1)}})
\label{eq:overhang_dist}
\end{equation}
To avoid collisions with trees, the maximum allowed distance $d_s^{max}$ is calculated as $d_s^{max}=(D-w)/2-d_e$, where $D$ is the row width and $d_e$ is a user-specified safety distance. The optimal covering circle is determined by continuing the iteration until the first occurrence where $d_s^i \leq d_s^{max}$ is met. The optimal circles, $\mathscr{C}^*(\vect{p}_c^*, r_c^*)$, are selected in this iteration, and the obstacles are inflated by $r_c^*$. Although iterating more would result in a smaller inflation, it would also require more circles to cover the vehicle, thus increasing the collision-checking time. The covering circles for the implement, denoted $\hat{\mathscr{C}}(\hat{\vect{p}}_c, \hat{r}_c)$, are computed based on $r_c^*$. To ensure no collision in the inflated space, $\hat{r}_c$ must be smaller than $r_c^*$. Therefore, $j$ increases incrementally until the first occurrence that satisfies $\hat{r}_c^j \leq r_c^*$, at which point the optimal covering circles for the implement are found.

Once all covering circles are determined, only the outermost circles for both the vehicle and implement are used for collision checking. For example, in Fig.~\ref{fig:circle_cover_and_max}, the optimal covering circles are found at $i=4$ and $j=3$. The obstacles are inflated by $r_c^*$, shown in gray. In hybrid A*, we only check whether the center points of the outermost circles collide with any obstacles in the inflated configuration space. Although hybrid A* generates collision-free trajectories, it cannot guarantee smoothness due to discretization and neglect of higher-order dynamic continuity. Therefore, the front-end trajectory must be refined in the back end to improve smoothness and optimality while ensuring compliance with operational constraints (for a detailed comparison, see~\cite{Peng2024}).

\subsection{Differential Flat Kinematic Model}
\label{sec:method_differential_flatness}
    
To reduce computational overhead in the back end, we leverage the differential flatness property of a simplified kinematic bicycle model to express vehicle dynamics in Cartesian space~\cite{Murray1995}. The flat output is defined as $\vect{\sigma} = [\sigma_x,\sigma_y]^T$, representing the vehicle's real axle center position $x$ and $y$. Using flat output and its finite derivatives, we can express arbitrary states of the vehicle at any given time. Denoting $\left\| \dot{\vect{\sigma}} \right\| = \sqrt{ \dot{\sigma}_x^2 + \dot{\sigma}_y^2}$, we derive the state of the vehicle as follows:
\begin{equation}
    \left\{\begin{aligned}
        x &= \sigma_x,~y = \sigma_y, v = \gamma \left\| \dot{\vect{\sigma}} \right\| \\
        \theta &= \arctan2(\gamma \dot{\sigma}_y,~\gamma \dot{\sigma}_x),~\dot{\theta} = \ddot{\vect{\sigma}}^T \vect{H} \dot{\vect{\sigma}} / \left\| \dot{\vect{\sigma}} \right\|^2 \\
        a &= \gamma \ddot{\vect{\sigma}}^T \dot{\vect{\sigma}} / \left\| \dot{\vect{\sigma}} \right\|,~\kappa = \gamma \ddot{\vect{\sigma}}^T \vect{H} \dot{\vect{\sigma}} / \left\| \dot{\vect{\sigma}} \right\|^3
    \end{aligned}
    \right.
\label{eq:differential_flatness}
\end{equation}
where $v$ is the linear velocity, $a$ is the acceleration, $\kappa$ is the curvature, and $\phi$ is the front steering angle. $\gamma\in \{-1, 1\}$ represents the motion direction of the AAV, and $L$ is the wheelbase length. $\vect{H}=[0, -1; 1, 0 ]$ is an auxiliary matrix.

\subsection{Nonlinear Optimization Problem Formulation}
\label{sec:method_optimization_problem}

Inspired by the \textit{MINCO} framework in \cite{Wang2022}, we formulate the vehicle trajectory $\mathcal{F}(t): \mathbb{R}^+ \mapsto \mathbb{R}^2$, represented by $\vect{\sigma}(t)$, as 2-dimensional polynomials of degree $2s-1$, where $s\in \mathbb{Z}^+$ is a user-specified parameter. The trajectory is divided into $N\in \mathbb{Z}^+$ segments, each with a constant motion direction. The $i$-th segment, for $i\in \{1,\ldots,N\}$, has a duration of $T_i\in \mathbb{R}^+$ and is further divided into $M_i\in \mathbb{Z}^+$ (with $M_i>1$) time-uniform pieces, each with a duration of $\tau_i = T_i / M_i$. Each piece $j\in\{1,\ldots,M_i\}$ within the $i$-th segment is parameterized by a coefficient matrix $\vect{c}_{i,j} \in \mathbb{R}^{2s \times 2}$ and a time basis vector $\vect{\beta}(t)\in \mathbb{R}^{2s}$ for $\forall t\in [0,~\tau_i]$:
\begin{equation}
\vect{\sigma}_{i,j}(t) = \vect{c}^T_{i,j} \vect{\beta}(t),~\vect{\beta}(t) = [t^0,t^1,\cdots, t^{2s-1}]^T
\label{eq:polynomial_trajectory}
\end{equation}

To effectively represent the problem, we define two coefficient matrices: $\vect{c}_i = [\vect{c}_{i,1}^T, \ldots, \vect{c}_{i,M_i}^T]^T$ and $\vect{\mathcal{C}} = [\vect{c}_1^T, \ldots, \vect{c}_N^T]^T $, along with a time vector $\vect{\mathcal{T}} = [T_1, \ldots, T_N]^T$. Additionally, we define the constraint function $\mathcal{G}_{\xi}$ for each $\xi\in \Xi$ as an inequality that depends on the flat output and its finite derivatives, $\mathcal{G}_{\xi}(\vect{\sigma}(t),\vect{\dot{\sigma}}(t),\ldots,\vect{\sigma}^{(s)}(t)) \preceq 0$, where $\Xi$ is the set of constraint terms, including dynamic feasibility constraints $\{v, a, \kappa, \dot{\theta} \}$, and the collision avoidance constraint $\varepsilon$. Let $\vect{\sigma}^{[d]} =[{\vect\sigma}^T, {\vect{\dot{\sigma}}}^T, \ldots, {\vect{\sigma}^{(d)}}^T]^T$, where $d\in \mathbb{Z}^+$. With these definitions, our problem can be formulated as a nonlinear constrained optimization problem that penalizes both control efforts and trajectory time:
\begin{subequations}
    \begin{align}
        \min_{\vect{\mathcal{C}}, \vect{\mathcal{T}}}~&J(\vect{\mathcal{C}}, \vect{\mathcal{T}}) = \int^{T_{\Sigma}}_{0} \vect{\sigma}^{(s)}(t)^T \vect{\sigma}^{(s)}(t)dt + w_T T_{\Sigma} \label{eq:original_cost}\\ 
        \text{s.t.}\quad & \vect{\sigma}_{1,1}^{[s-1]}(0) = \vect{S}_0, ~\vect{\sigma}_{N,M_N}^{[s-1]}(T_{\Sigma}) = \vect{S}_N \label{eq:start_end_condition}\\ 
        & \vect{\sigma}_{i,M_i}^{[s-1]}(T_i) \!=\! \vect{\sigma}_{i+1,1}^{[s-1]}(0) \!=\! \vect{S}_i,\forall i \!\in\! \!\{1, \!\ldots\!, N\!-\!1\} \label{eq:segment_condition}\\ 
        & \begin{aligned}[t]
        & \vect{\sigma}_{i,j}^{[2(s-1)]}(\tau_i) = \vect{\sigma}_{i, j+1}^{[2(s-1)]}(0), ~\forall i \!\in\! \{1, \!\ldots\!, N\}, \\
        & \forall j \!\in\! \{1, \!\ldots\!, M_i\!-\!1\}\end{aligned} \label{eq:piece_condition}\\ 
        & \mathcal{G}_{\xi}(\vect{\sigma}(t), \!\ldots\!, \vect{\sigma}^{(s)}(t)) \preceq 0,~\forall \xi \!\in\! \Xi,~\forall t \!\in\! [0,T_{\Sigma}] \label{eq:inequality_constraint}\\
        & 0 < T_i < T_{i+1},~\forall i \!\in\! \{1, \!\ldots\!, N\!-\!1\} \label{eq:postiveness} 
    \end{align}
    \label{eq:constrained_optimization_problem}
\end{subequations} 

\noindent where $T_{\Sigma} = \sum_{i=1}^{N} T_i$ is the total trajectory duration and $w_T$ is the weight on the time. $\vect{S}_0$ and $\vect{S}_N$ are the start and end conditions, respectively. $\vect{S}_i$ is the condition between two adjacent segments. In this work, we select $s=3$ to minimize the total jerk of the trajectory. 

\subsection{Relaxation of Constraints}
\label{sec:method_constraint_relaxation}

\eqref{eq:start_end_condition} and \eqref{eq:piece_condition} are inherently satisfied within the MINCO framework. The segment-adjacent point constraints in \eqref{eq:segment_condition} are relaxed using the GSPO algorithm from \cite{Han2023}, where the positions $\hat{p}$ and orientations $\hat{\theta}$ of adjacent points are treated as optimizable variables. This approach helps mitigate suboptimal solutions caused by the front-end trajectory. To satisfy the inequality constraints in \eqref{eq:inequality_constraint}, we discretize the $j$-th piece into $K_j$ constraint points and enforce constraints at these points. The integral of the continuous constraint function violations is then approximated by summing the penalties at the discretized constraint points:
\begin{subequations}
    \begin{align}
        &\mathcal{P}_{\Sigma}(\vect{\vect{\mathcal{C}}},\vect{\vect{\mathcal{T}}}) = \sum_{\xi\in \Xi} w_{\xi} \sum_{i=1}^N \sum_{j=1}^{M_i} \sum_{\lambda=0}^{K_j} \mathcal{P}_{\xi,i,j,\lambda}(\vect{c}_{i,j}, T_i) \\
        &\mathcal{P}_{\xi,i,j,\lambda}(\vect{c}_{i,j}, T_i) = \bar{w}_{\lambda} \alpha_{i,j} \mathcal{L}_1(\mathcal{G}_{\xi,i,j,\lambda})
    \end{align}
\end{subequations}
where $\alpha_{i,j}=\tau_i/K_j$ and $\mathcal{L}_1(\cdot)$ is the L1-norm relaxation function. $w_{\xi}$ is the penalty weight corresponding to constraint $\xi$, and $\bar{w}_{\lambda}$ are predetermined quadrature coefficients. According to the chain rule, the gradients of the constraint penalty with respect to $\vect{c}_{i,j}$ and $T_i$ are derived as:
\begin{subequations}
    \begin{align}
        \frac{\partial \mathcal{P}_{\xi,i,j,\lambda}}{\partial \vect{c}_{i,j}} &= \bar{w}_{\lambda} \alpha_{i,j} \dot{\mathcal{L}}_1(\mathcal{G}_{\xi,i,j,\lambda}) \bigg[ \sum_{d=0}^{s} \beta^{(d)}(\bar{t}) \Big( \frac{\partial \mathcal{G}_{\xi,i,j,\lambda}}{\partial \vect{\sigma}^{(d)}_{i,j}} \Big)^T \bigg] \label{eq:penalty_gradient_a} \\ 
        \frac{\partial \mathcal{P}_{\xi,i,j,\lambda}}{\partial T_i} &= \frac{\mathcal{P}_{\xi,i,j,\lambda}}{T_i} + \bar{w}_{\lambda} \alpha_{i,j} \dot{\mathcal{L}}_1(\mathcal{G}_{\xi,i,j,\lambda}) \bigg[\frac{\lambda}{M_i K_j}\cdot \\
        \nonumber &\qquad \sum_{d=0}^{s} {\vect{\sigma}^{(d+1)}_{i,j,\lambda}}^T \Big( \frac{\partial \mathcal{G}_{\xi,i,j,\lambda}}{\partial \vect{\sigma}^{(d)}_{i,j}} \Big) \bigg]
    \end{align}
\label{eq:penalty_gradients}
\end{subequations}

The constraint function violations and their gradients can be computed using the differential flatness relations in \eqref{eq:differential_flatness}. For example, if the maximum allowed angular velocity is $\dot{\theta}_{max}$, the corresponding constraint violation is:
\begin{equation}
    \mathcal{G}_{\dot{\theta}}(\dot{\vect{\sigma}}, \ddot{\vect{\sigma}}) = \big( \frac{\ddot{\vect{\sigma}}^T \vect{H} \dot{\vect{\sigma}}} {\left\| \dot{\vect{\sigma}} \right\|^2} \big)^2 - \dot{\theta}^2_{max}
    \label{eq:angular_velotity_violation}
\end{equation}
with the gradients:
\begin{subequations}
    \begin{align}
    \frac{\partial \mathcal{G}_{\dot{\theta}}}{\partial \dot{\vect{\sigma}}} &= 2\frac{\ddot{\vect{\sigma}}^T \vect{H} \dot{\vect{\sigma}}}{\left\| \dot{\vect{\sigma}} \right\|^4}  \vect{H}^T \ddot{\vect{\sigma}} - 4\frac{(\ddot{\vect{\sigma}}^T \vect{H} \dot{\vect{\sigma}})^2} {\left\| \dot{\vect{\sigma}} \right\|^6}  \dot{\vect{\sigma}} \\
    \frac{\partial \mathcal{G}_{\dot{\theta}}}{\partial \ddot{\vect{\sigma}}} &= 2 \frac{\ddot{\vect{\sigma}}^T \vect{H} \dot{\vect{\sigma}}}{\left\| \dot{\vect{\sigma}} \right\|^4}  \vect{H} \dot{\vect{\sigma}}
    \end{align}
    \label{eq:angular_velocity_limitation}
\end{subequations}

The discretized gradient at each constraint point can be calculated using \eqref{eq:penalty_gradients}. Other constraints and their derivatives are defined similarly to those in \cite{Han2023} and are omitted here for brevity. Additionally, the positive clock time $\vect{\mathcal{T}}$ is remapped to unbounded virtual time $\tilde{\vect{\mathcal{T}}}$ to relax the constraint in \eqref{eq:postiveness}.

\begin{figure}[t]
\centering 
    \includegraphics[width=0.42\columnwidth]{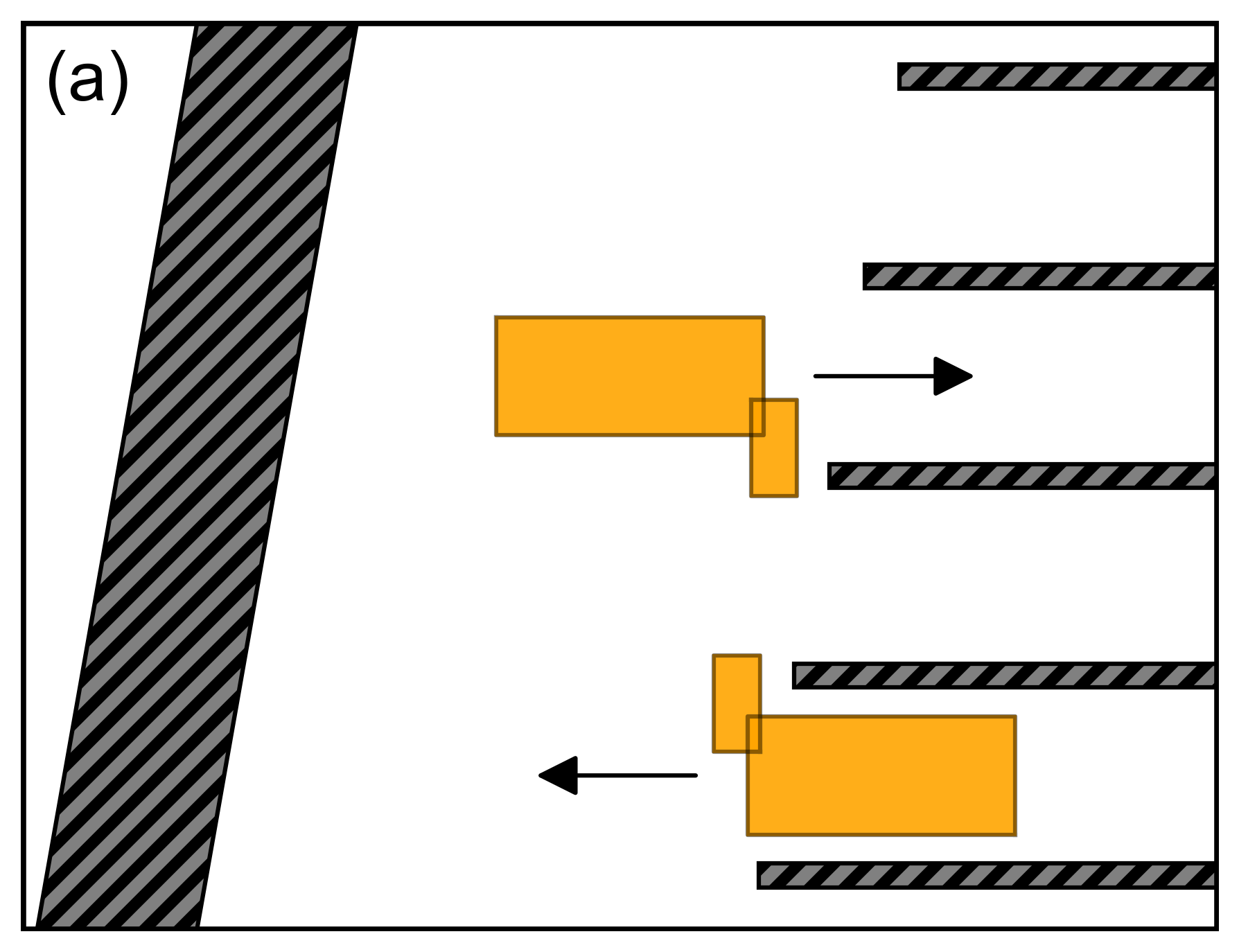}
    \includegraphics[width=0.42\columnwidth]{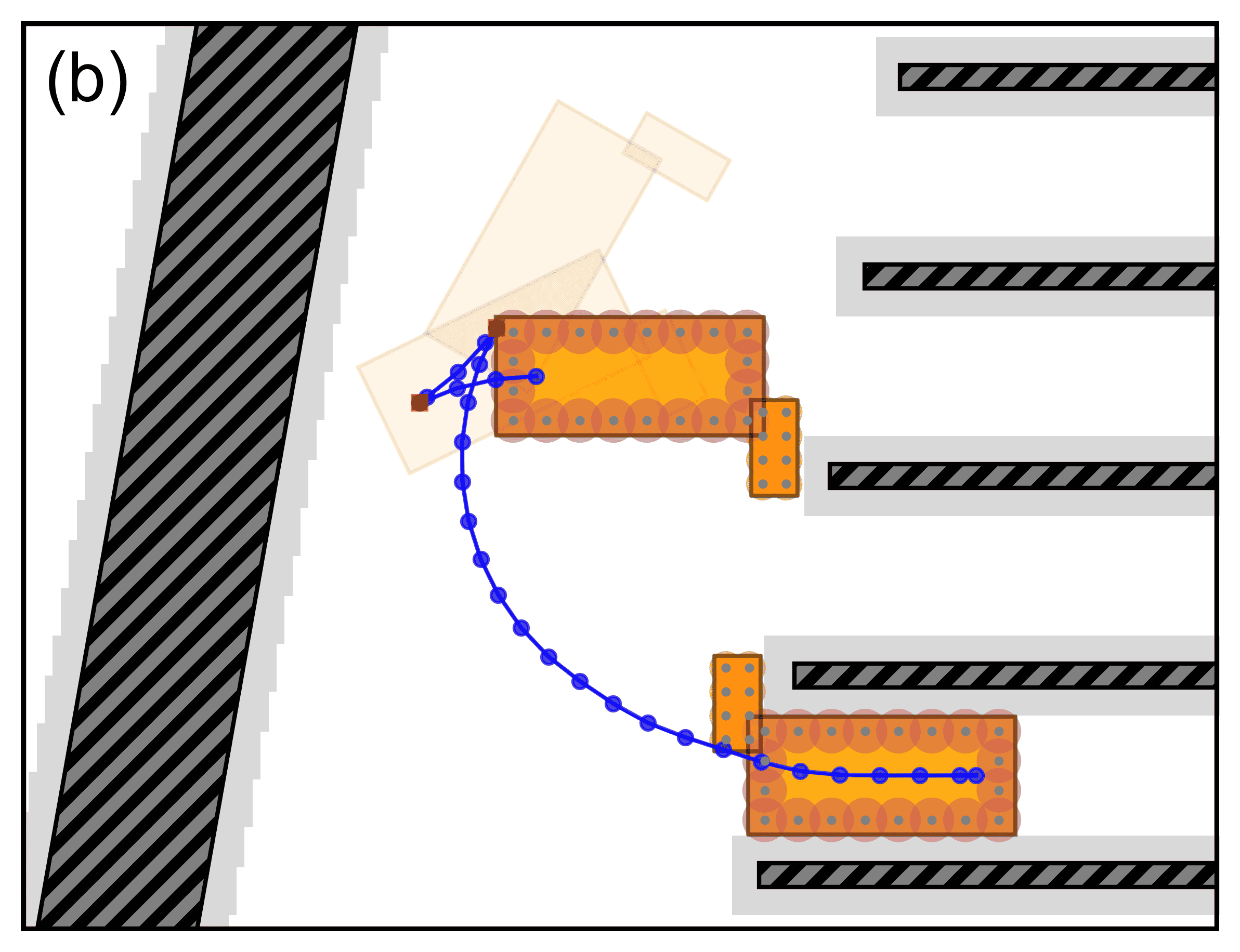}
    \includegraphics[width=0.42\columnwidth]{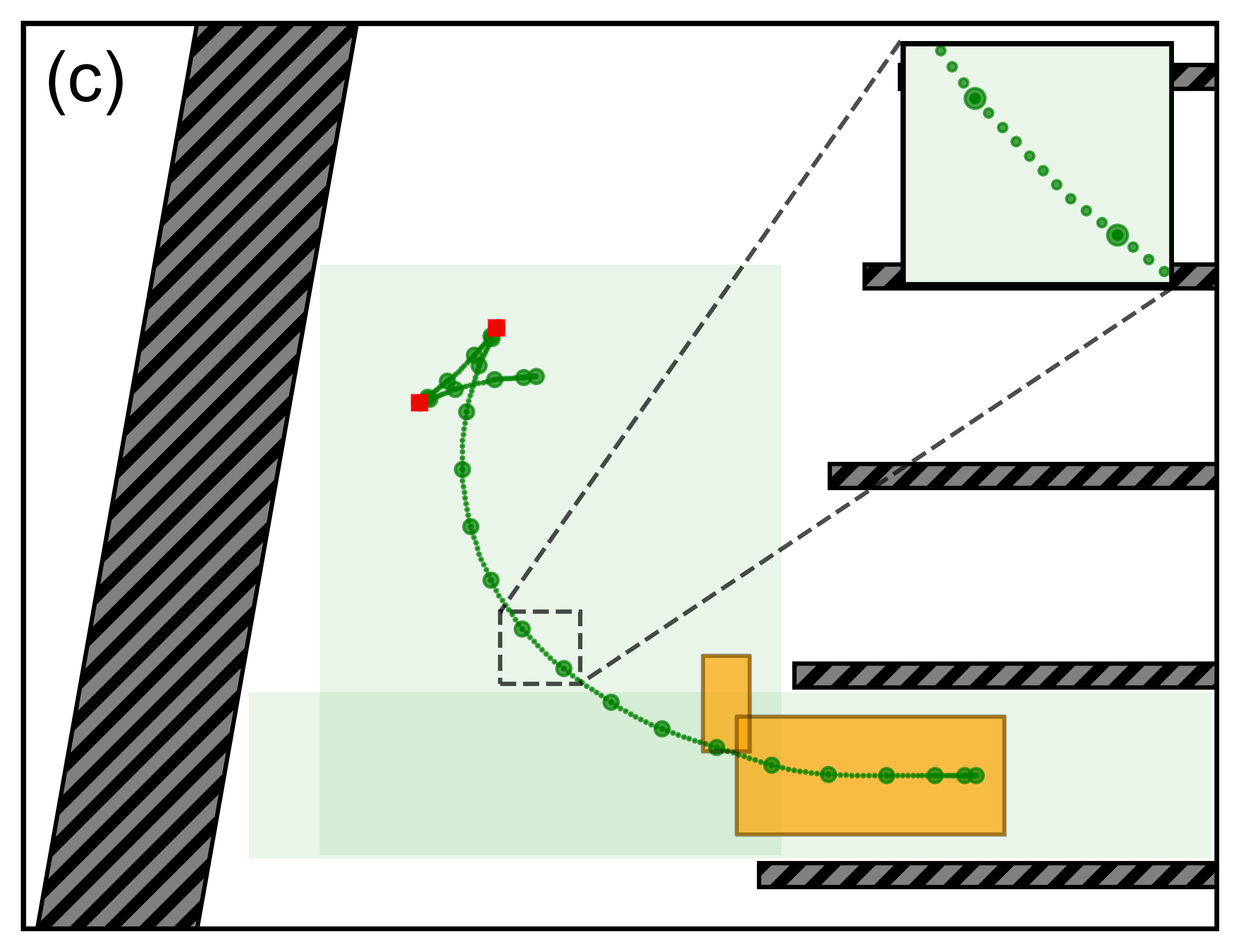}
    \includegraphics[width=0.42\columnwidth]{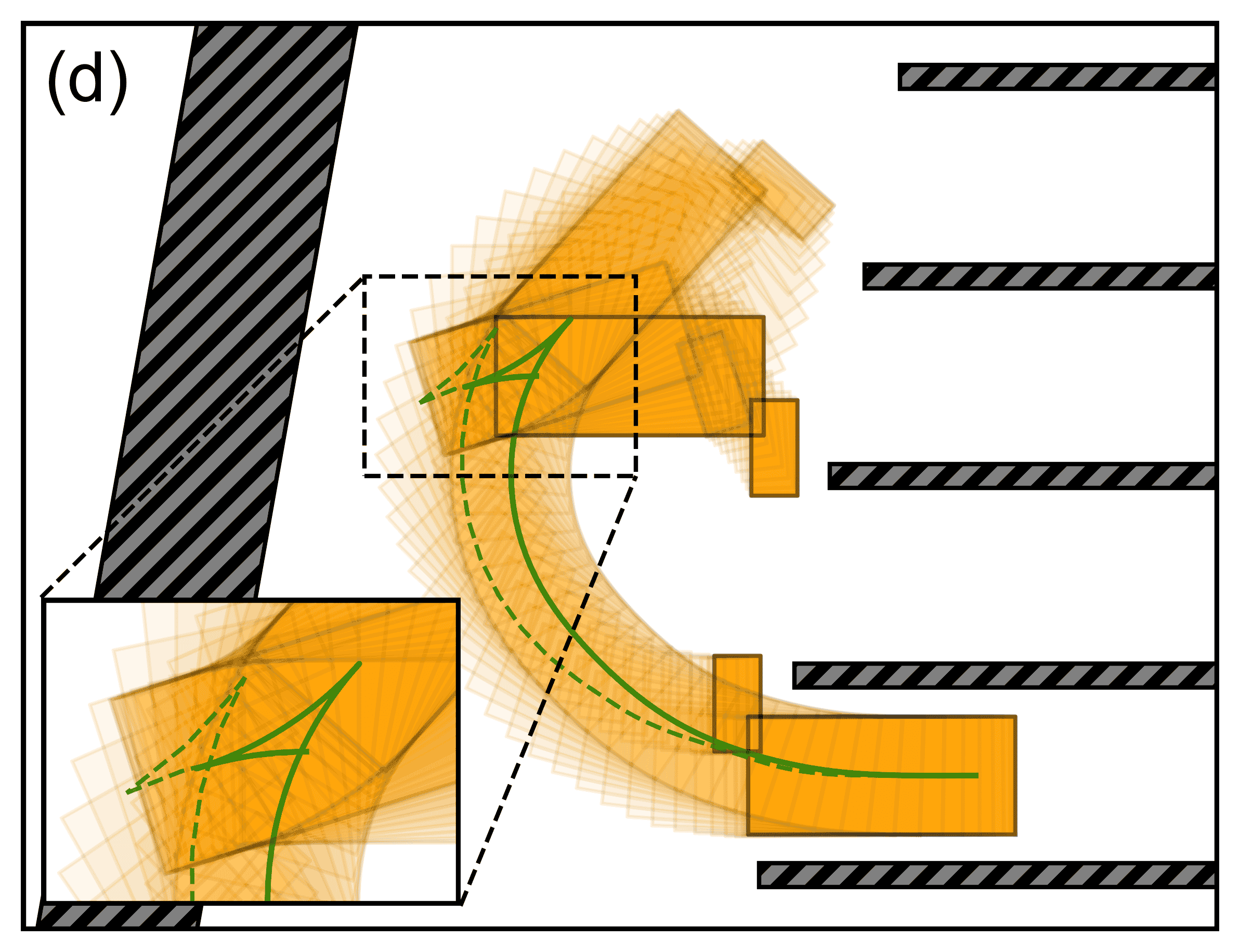}
    \includegraphics[width=0.9\columnwidth]{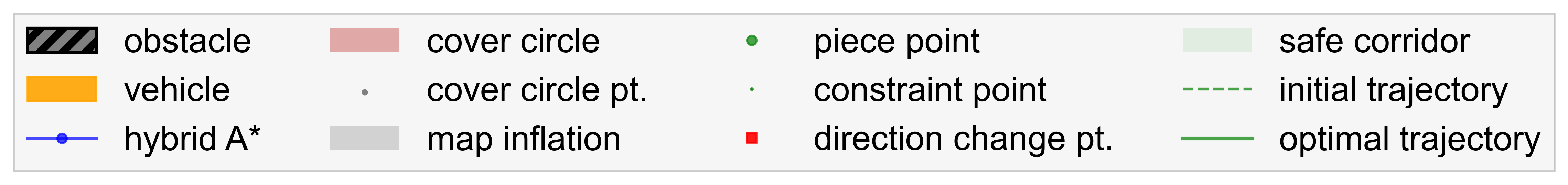}
    \caption{Illustration of (a) geometric representations of obstacles and the AAV; (b) results from the front-end search using covering circles; (c) the parameterized initial trajectory with multiple segments, piece points, and constraint points; and (d) the optimized trajectory following the back-end refinement.}
\label{fig:algorithm}
\end{figure}

\subsection{Corridor-based Obstacle Avoidance}
\label{sec:method_safe_corridor}

Both the vehicle and its implement must avoid collisions during headland turning. Achieving whole-body obstacle avoidance in unstructured environments is challenging, particularly when the vehicle carries implements with complex shapes~\cite{Geng2023}. To improve efficiency, we construct convex safe corridors to keep the AAV within these regions and reduce search space during optimization. Although enclosing both the vehicle and the implement in the same corridor is possible, doing so is often overly conservative and can lead to infeasible solutions in narrow spaces. To overcome this, we compute separate convex corridors for the vehicle and implement, which allows for more flexible and efficient planning.

We enforce the obstacle avoidance constraint $\varepsilon$ at each constraint point. Using the environmental map and the front-end result, we calculate rectangular corridors with maximal safe areas for each part of the AAV at these points.  The corridors are computed once after obtaining the initial trajectory from the front end and remain fixed during optimization. Each corridor aligns with the vehicle's orientation at its respective constraint point and is formed by expanding the edges of each part outward until they intersect obstacles. Let $k$ represent a combination of $i,j,\lambda$. The H-representation of the safe corridor at time step $t_k$ is defined as $SC^H_{\zeta,k} = \{\vect{x}\in\mathbb{R}^2: \vect{A}_{\zeta,k}\vect{x} \leq \vect{b}_{\zeta,k} \}$, where $\vect{A}_{\zeta,k} = [\vect{A}_{\zeta,k}^1, \ldots, \vect{A}_{\zeta,k}^{N_{\zeta,k}}]^T$ and $\vect{b}_{\zeta,k} = [b_{\zeta,k}^1, \ldots, b_{\zeta,k}^ {N_{\zeta,k}}]^T$. Here, $\zeta \in \{\zeta_{veh,1},\ldots,\zeta_{veh,N_{vel}},\zeta_{imp,1},\ldots,\zeta_{imp,N_{imp}}\}$ denotes the individual parts of the vehicle and implement, and $N_{\zeta,k}$ is the number of hyperplanes. Collision avoidance is ensured by enforcing $\mathbb{V}_m(t_k) \subseteq SC^H_{\zeta_{veh,m},k}$ for $m\leq N_{veh}$ and $\mathbb{V}_n(t_k) \subseteq SC^H_{\zeta_{imp,n},k}$ for $n\leq N_{imp}$. Since the vehicle and implement are rigidly connected, the vertex sets of the vehicle and implement rectangles can be derived as:
\begin{equation}
    \mathcal{V}_{\zeta} = \big\{\vect{v}_{\zeta, e}\in \mathbb{R}^2: \vect{v}_{\zeta, e} = \vect{\sigma} + \vect{R} \vect{V}_{\zeta, e},~\forall e\in \{1, \ldots, n_{\zeta}\} \big\}
\end{equation}
where $\vect{V}_{\zeta, e}$ are the rectangle vertices in the body frame, $n_{\zeta}=4$ is the number of vertices, and $\vect{R}$ is the rotation matrix of the AAV. Ensuring that the full shape of the vehicle or implement is contained within a safe corridor is done by enforcing that all vertices lie within the corresponding convex polygon. The collision avoidance constraint violation at any point is computed as:
\begin{equation}
    \mathcal{G}_{\zeta, k, e}(\dot{\vect{\sigma}}, \ddot{\vect{\sigma}}) = \vect{A}^T_{\zeta, k}(\vect{\sigma} + \vect{R}\vect{V}_{\zeta, e})- \vect{b}_{\zeta, k}
\end{equation}
with its gradients derived in \cite{Han2023}. Note that while the front-end algorithm searches for a feasible initial trajectory in a grid space, the back-end algorithm performs optimization in a continuous space bounded by the combinations of safe corridors.

\subsection{Solve an Unconstrained Optimization Problem}
\label{section:method_problem_reformulation}
The original optimization problem in \eqref{eq:constrained_optimization_problem} is converted into an unconstrained nonlinear optimization problem. Using the MINCO, the coefficient matrix $\vect{\mathcal{C}}$ can be determined by the intermediate waypoints $\vect{q}$, unbounded time variables $\tilde{\vect{\mathcal{T}}}$, and the positions \(\hat{\vect{p}}\) and orientations \(\hat{\vect{\theta}}\) of the segment-adjacent points. Therefore, we are now solving the following unconstrained optimization problem:
\begin{equation}
    \min_{\vect{q}, \hat{\vect{p}}, \hat{\vect{\theta}}, \tilde{\vect{\mathcal{T}}}} \mathcal{J}(\vect{q}, \hat{\vect{p}}, \hat{\vect{\theta}}, \tilde{\vect{\mathcal{T}}}) = \min_{\vect{\mathcal{C}}(\vect{q}, \hat{\vect{p}}, \hat{\vect{\theta}} , \tilde{\vect{\mathcal{T}}}), \tilde{\vect{\mathcal{T}}}} J(\vect{\mathcal{C}}, \tilde{\vect{\mathcal{T}}}) + \mathcal{P}_{\Sigma}(\vect{\mathcal{C}}, \tilde{\vect{\mathcal{T}}})
\label{eq:unconstrained_optimization}
\end{equation}

In this reformulation, all constraints from the original optimization have been relaxed and removed. The gradients with respect to different variables are derived analytically, and the unconstrained optimization is solved using the L-BFGS optimizer~\cite{Liu1989}. Fig.~\ref{fig:algorithm} illustrates the main steps in our trajectory planning process.

\section{Experiments and Results}
\label{sec::experiments_and_results}

We conducted a series of experiments to evaluate the performance of our trajectory planner. The experiments were divided into two sets. The first set was performed in simulations, using an AAV model based on a commercial autonomous electric tractor by \textit{VitalTech Co.}~\cite{VitalTech}. The second set tested the performance on a physical tricycle robot~\cite{Peng2024}, equipped with a spray boom, operating in a vineyard on the UC Davis University campus. The parameters for both AAVs were configured in the planner according to Table~\ref{tab:vehicle_parameters}.

\begin{table}[t]
\centering
\caption{AAV parameters used in simulations and physical robot experiments.}
    \begin{tabular}{lcc}
    \toprule
    \textbf{Parameter} & \textbf{Value (Sim.)} & \textbf{Value (Phys. robot)} \\
    \midrule
    Vehicle width (m)                   & 1.48 & 1.2 \\
    Vehicle length (m)                  & 3.35 & 1.8 \\
    Wheelbase (m)                       & 1.9 & 1.285 \\
    Max. linear velocity (m/s)          & 1.5 & 1.0 \\
    Max. acceleration (m/s$^2$)         & 1.0 & 0.6 \\
    Max. curvature (m$^{-1}$)           & 0.323 &  0.323 \\
    Max. angular velocity (rad/s)       & 0.5 & 0.5 \\
    \bottomrule
    \end{tabular}
\label{tab:vehicle_parameters}
\end{table}

\subsection{Results in Simulations}
\label{sec:results_simulations}

In simulation experiments, we evaluated our planner for an AAV equipped with four types of implements (see Fig.~\ref{fig:tractor_implement}). The planner was assessed in two scenarios: (1) \textit{standard} orchards (see Fig.~\ref{fig:algorithm}(a) as an example) with varying headland widths between 6.5 m and 8.0 m, and (2) \textit{non-standard} orchards with irregular field boundaries and crop row ends (the headland width is roughly 6.0-7.0 m) to mimic some real-world complexity. The AAV was tasked to transit between two non-adjacent rows with predefined start and end poses. Simulations were conducted on an Intel i7-13700F processor with 16GB of memory.  

\begin{table*}[t]
\centering
\caption{Comparison of the front-end and back-end algorithms in standard and non-standard orchards.}
\def\rowheight{2.5ex}
\begin{NiceTabular}{c|c|rrrrr||c|rrr}
    \toprule
    \multirow{3}{*}{\bf{\makecell[l]{Impl. \\ type}}} & \multicolumn{6}{c}{\bf{Computation Time in Standard Orchards}} & \multicolumn{4}{c}{\bf{Computation Time in Non-standard Orchards}} \\
    \cline{2-7} \cline{8-11} \rule{0pt}{\rowheight}
    & \multirow{2}{*}{\bf{\makecell[c]{Headland \\ width (m)}}} & \multicolumn{2}{c}{\bf{front-end time (ms)}} & \multicolumn{3}{c}{\bf{back-end time (Traj. duration)} (s)} & \multirow{2}{*}{\bf{Case}} & \multicolumn{3}{c}{\bf{back-end time (Traj. duration)} (s)} \\
    \cline{3-4} \cline{5-7} \cline{9-11} \rule{0pt}{\rowheight}
    & & \bf{Raycast} & \bf{Proposed (diff.)} & \bf{OBCA} & \bf{SRA} & \bf{Proposed} & & \bf{OBCA} & \bf{SRA} & \bf{Proposed} \\
    \cline{1-11} \rule{0pt}{\rowheight}
    \multirow{4}{*}{\makecell[l]{Double \\ pruner}} & 6.5 & 19.5 & 5.5 (-71.8\%) & 8.4 (13.9) & 1.9 (21.8) & 1.3 (14.5) & I & 15.0 (12.5) & 1.4 (23.2) & 1.1 (13.0)\\
                                                    & 7.0 & 18.8 & 3.1 (-83.5\%) & 8.3 (14.2) & 1.3 (20.1) & 1.6 (13.8) & II & 11.2 (13.8) & 0.7 (16.6) & 0.9 (14.1) \\
                                                    & 7.5 & 22.2 & 3.8 (-82.9\%) & 6.5 (16.1) & 0.9 (16.9) & 1.7 (13.5) & III & 41.1 (14.1) & 1.7 (23.2) & 1.1 (14.2) \\
                                                    & 8.0 & 13.5 & 3.2 (-76.3\%) & 6.2 (14.2) & 0.7 (17.1) & 1.2 (13.5) & IV & fail & 0.8 (16.2) & 1.0 (14.0) \\
    \cline{1-11} \rule{0pt}{\rowheight}
    \multirow{4}{*}{\makecell[l]{Single \\ pruner}} & 6.5 & 6.4 & 4.7 (-26.6\%) & 6.5 (13.8) & 0.6 (16.1) & 1.6 (14.6) & I & 9.9 (14.0) & 0.4 (16.6) & 1.0 (13.9) \\
                                                    & 7.0 & 6.8 & 2.0 (-70.6\%) & 15.1 (14.2) & 0.6 (16.3) & 1.4 (13.9) & II & fail & 0.6 (16.4) & 0.7 (14.4) \\
                                                    & 7.5 & 7.9 & 5.3 (-32.9\%) & 4.6 (14.5) & 0.5 (16.5) & 1.0 (14.1) & III & 14.9 (13.6) & 1.8 (17.7) & 1.5 (13.2) \\
                                                    & 8.0 & 8.5 & 2.6 (-60.4\%) & 4.8 (14.4) & 0.5 (16.3) & 1.2 (14.3) & IV & 9.7 (13.2) & 0.6 (15.8) & 1.2 (13.1) \\
    \cline{1-11} \rule{0pt}{\rowheight}
    \multirow{4}{*}{Mower}                          & 6.5 & 260.6 & 37.4 (-85.6\%) & 8.4 (17.5) & 1.1 (20.2) & 1.5 (16.8) & I & 19.1 (19.8) & 0.5 (22.0) & 1.3 (20.0) \\
                                                    & 7.0 & 188.7 & 55.0 (-70.1\%) & 9.1 (17.8) & 0.5 (18.1) & 1.2 (15.7) & II & 24.0 (18.7) & 0.3 (17.5) & 1.2 (18.9) \\
                                                    & 7.5 & 152.1 & 73.2 (-51.8\%) & 16.4 (17.7) & 0.6 (17.4) & 0.9 (16.1) & III & fail & 1.0 (20.5) & 1.2 (20.2) \\
                                                    & 8.0 & 189.9 & 56.0 (-70.5\%) & 7.7 (17.3) & 0.6 (16.8) & 0.9 (14.3) & IV & fail & 0.7 (16.9) & 1.5 (19.4) \\
    \cline{1-11} \rule{0pt}{\rowheight}
    \multirow{4}{*}{\makecell[l]{KMS \\ sprayer}}   & 6.5 & fail & fail & fail & fail & fail & I & 35.0 (17.9) & fail & 1.5 (18.1) \\
                                                    & 7.0 & 419.2 & 22.1 (-94.7\%) & fail & 3.2 (25.9) & 1.6 (17.5) & II & fail & 33.4 (47.5) & 3.1 (23.8) \\
                                                    & 7.5 & 883.6 & 176.5 (-80.0\%) & 28.6 (15.9) & 3.8 (30.7) & 1.0 (15.2) & III & 16.1 (18.6) & 6.3 (27.4) & 1.9 (17.0) \\
                                                    & 8.0 & 3123.5 & 368.3 (-88.2\%) & 7.5 (16.4) & 0.7 (17.8) & 1.3 (17.2) & IV & 16.3 (17.8) & 11.2 (36.9) & 1.1 (18.2) \\
    \cline{1-11} 
    \multicolumn{2}{l}{\rule{0pt}{\rowheight} \bf{Average:}}        & 354.8 & 54.6 (-84.6\%) & 15.5 (15.6) & 1.2 (19.2) & 1.3 (15.0) & & 21.7 (15.9) & 4.1 (22.3) & 2.1 (16.6) \\
    \multicolumn{2}{l}{\bf{Success rate:}}   & 15/16 & 15/16 (+0.0\%) & 14/16 & 15/16 & 15/16 & & 11/16 & 15/16 & 16/16 \\
    \bottomrule 
\end{NiceTabular}
\label{tab:comparison_results}
\end{table*}

We first compared our covering-circle collision detection method to the ray-casting algorithm \cite{Ericson2004} in the front-end search. For a fair comparison, all other aspects of the front-end process and the environment map remained the same, with only the collision detection function being swapped. The results in Table~\ref{tab:comparison_results} show that our method reduces search time by an average of 84.6\% in standard orchards. Notably, both methods failed at a headland width of 6.5 m when carrying the KMS sprayer due to limited space.

\begin{figure}[b!]
\centering
    \begin{subfigure}{0.42\linewidth}
        \includegraphics[width=\textwidth]{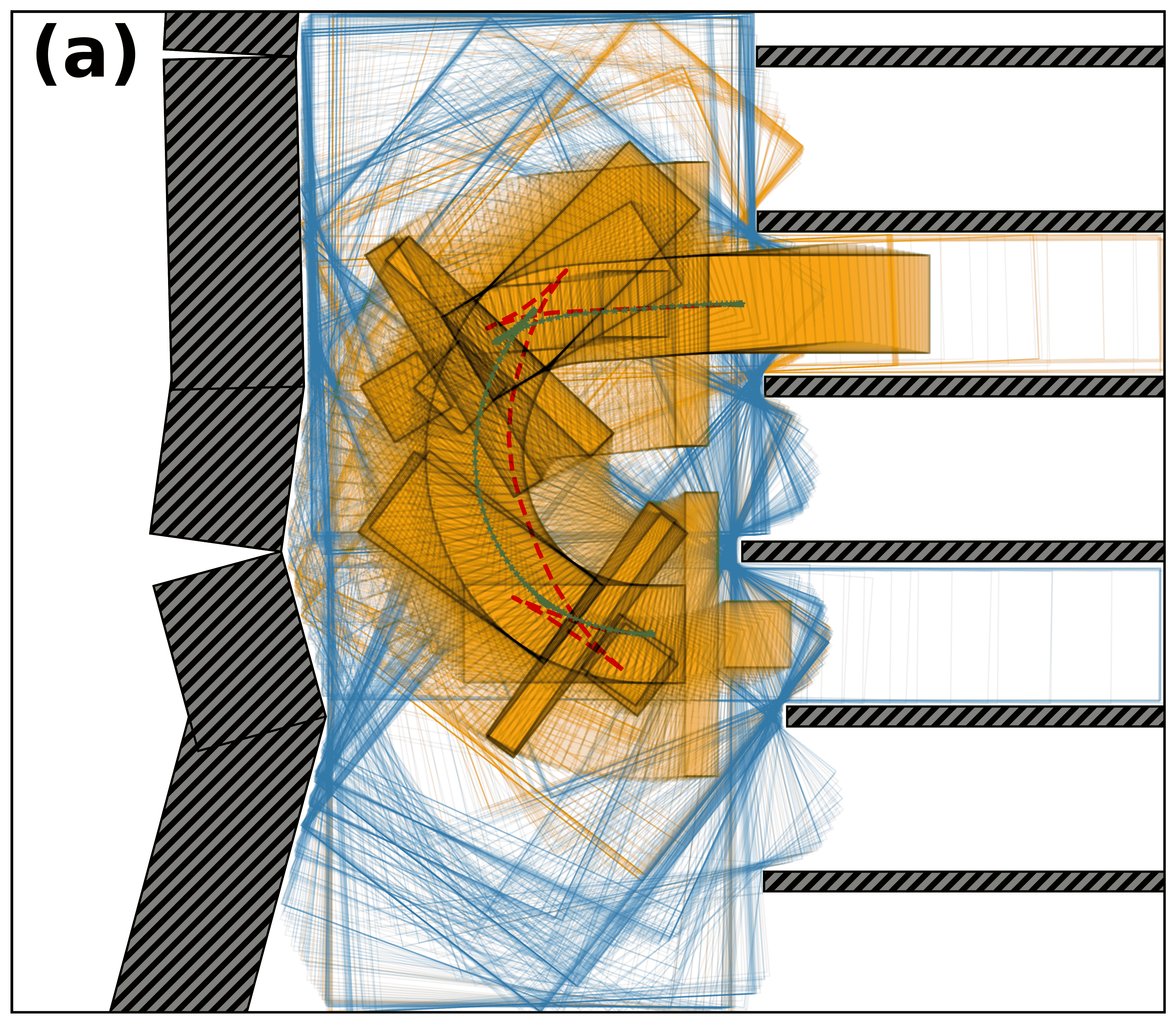}
    \end{subfigure}
    \begin{subfigure}{0.42\linewidth}
        \includegraphics[width=\textwidth]{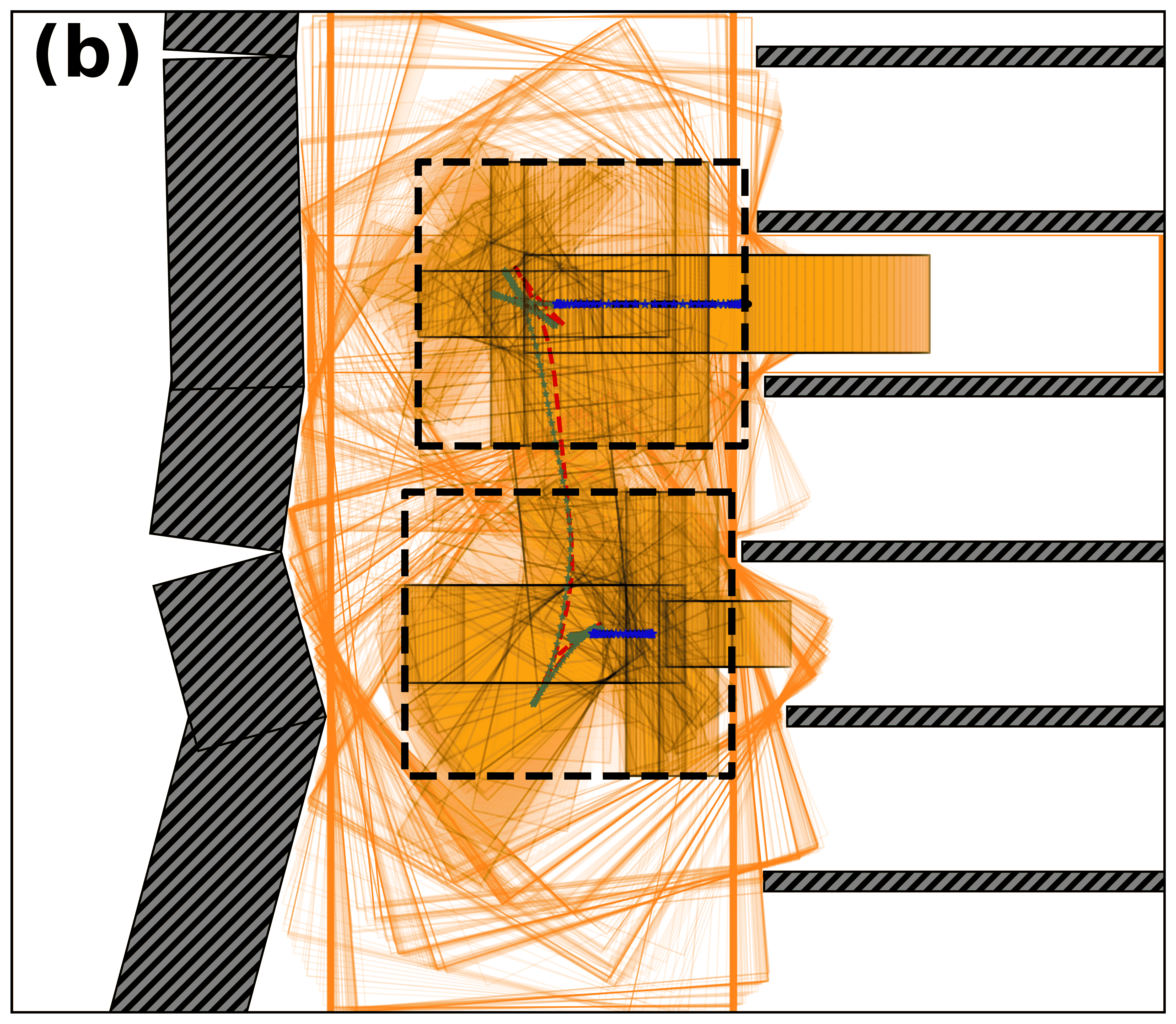}
    \end{subfigure}
    \begin{subfigure}{0.99\linewidth}
        \includegraphics[width=\textwidth]{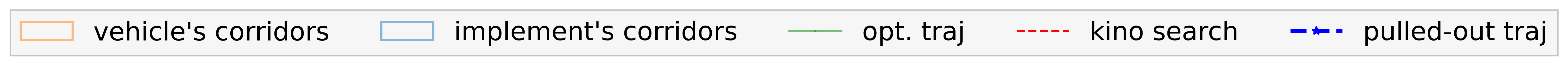}
    \end{subfigure}
    \caption{Comparison between (a) our planner and (b) SRA results carrying a KMS sprayer in Case II. The red dashed line shows the front-end trajectory and the green line shows the optimized trajectory from the back end. Safe corridors for both the vehicle and implement are depicted. The covering rectangle in SRA is shown with dashed black lines, and the dashed blue lines represent the pull-out trajectories.}
\label{fig:nonstandard_cases}
\end{figure}

Next, we compared our back-end optimizer with the OBCA method \cite{Peng2024} and Han's method~\cite{Han2023}. Our planner and OBCA used the same front-end results. In the original implementation of \cite{Han2023}, the vehicle body is represented as a single rectangle (see Fig.~\ref{fig:nonstandard_cases}(b) for an illustration), leading to immediate failure due to its conservatism at the designated start and end poses. To enable a fair comparison, we adjusted the start and end poses by pulling them away from the crop rows until the covering rectangle no longer intersected with obstacles. A new initial trajectory was then calculated based on the adjusted poses before applying Han’s method. The optimized trajectory was connected to the two pull-out segments using straight lines, resulting in the final trajectory. In our experiment, we denoted this adapted version of Han's method as the single-rectangle algorithm (SRA).

\begin{figure}[t]
\centering
    \includegraphics[width=0.97\linewidth]{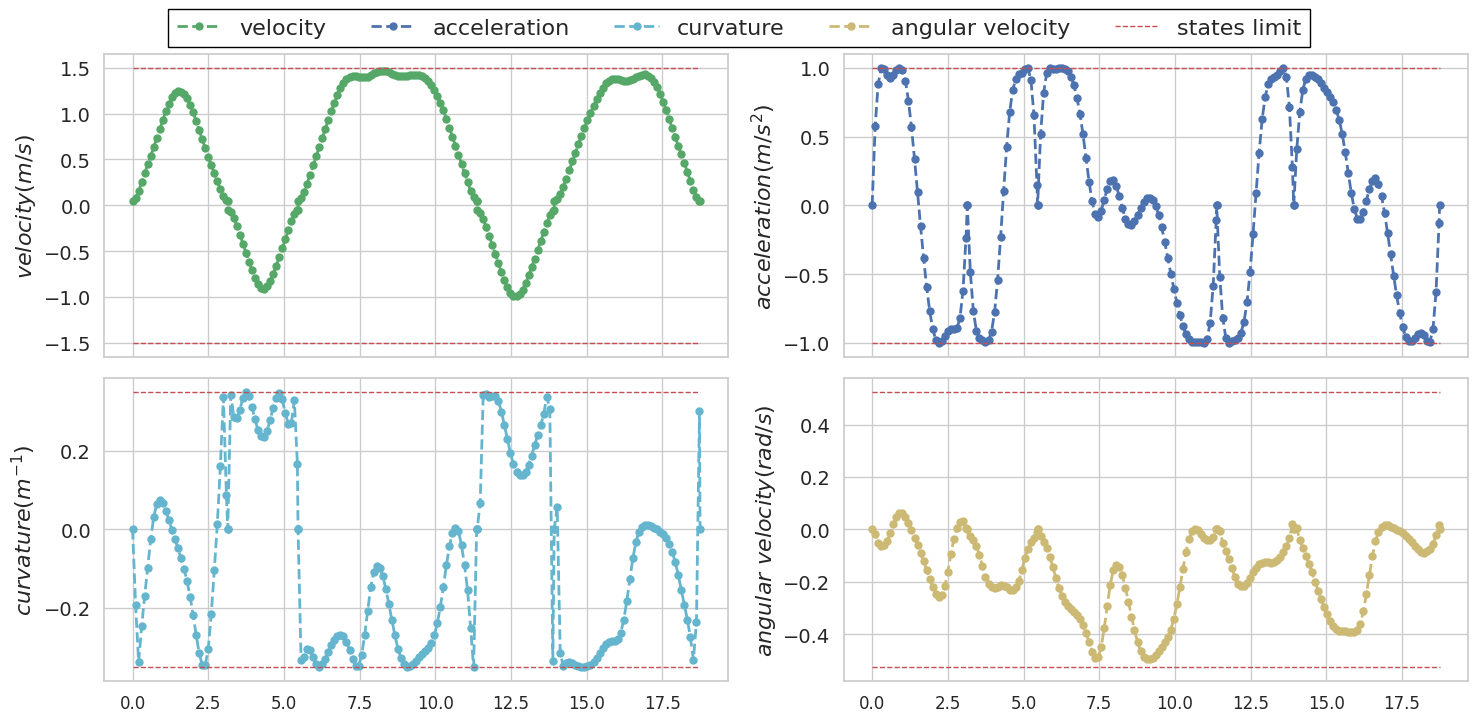}
    \caption{AAV states over time for the optimal trajectory with a KMS sprayer in Case IV.}
\label{fig:states_over_time}
\end{figure}

We provide the computation time and optimized trajectory durations for three back-end algorithms in Table~\ref{tab:comparison_results}. The results show that our planner requires 91.0\% less computation time than OBCA while achieving similar trajectory durations for successful cases. OBCA failed in five scenarios for non-standard cases, whereas our method succeeded in all cases. The SRA performed comparably to our method with shorter computation times in some cases due to fewer safe corridor constraints. However, its trajectory durations were longer than our planner's results. The SRA's performance dropped significantly when applied to the KMS sprayer and in non-standard orchards, where both the computation times and trajectory durations were substantially higher than our results and even failed in one scenario. To explain this difference, we plot the optimal trajectories generated by our planner and the SRA of one scenario in Fig.~\ref{fig:nonstandard_cases}. It can be observed that our optimized trajectory includes fewer direction changes and fully utilizes the available space, where SRA sacrifices due to its conservatism, leading to more complicated maneuvers in limited space. To demonstrate the optimality of our back-end optimizer, we depicted the AAV states over time in Fig.~\ref{fig:states_over_time}. The results show that our planner maintains all states within the predefined limits while ensuring smoothness.

\begin{figure}[t!]
\centering
    \begin{subfigure}{0.42\linewidth}
        \includegraphics[width=\textwidth]{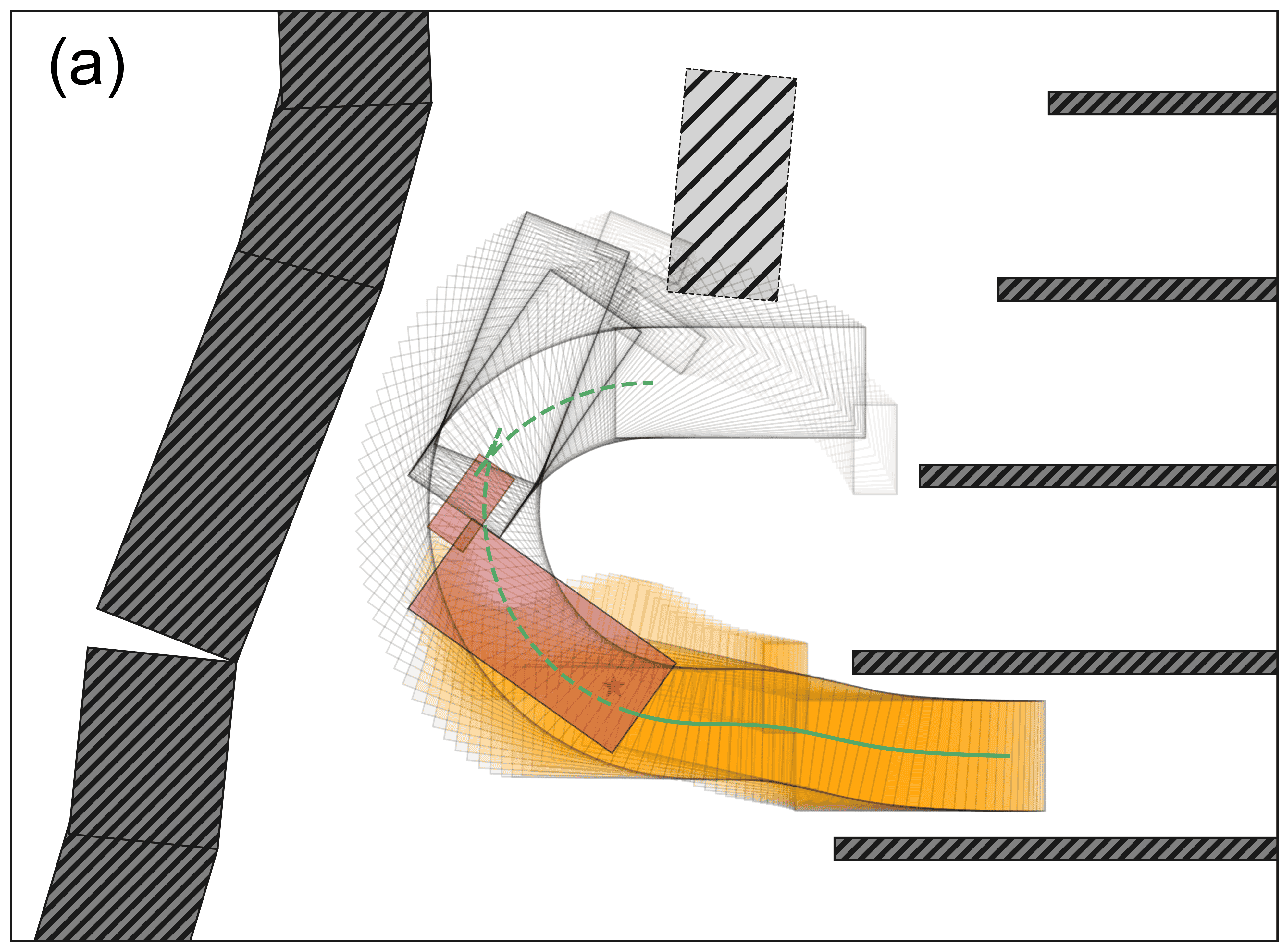}
    \end{subfigure}
    \begin{subfigure}{0.42\linewidth}
        \includegraphics[width=\textwidth]{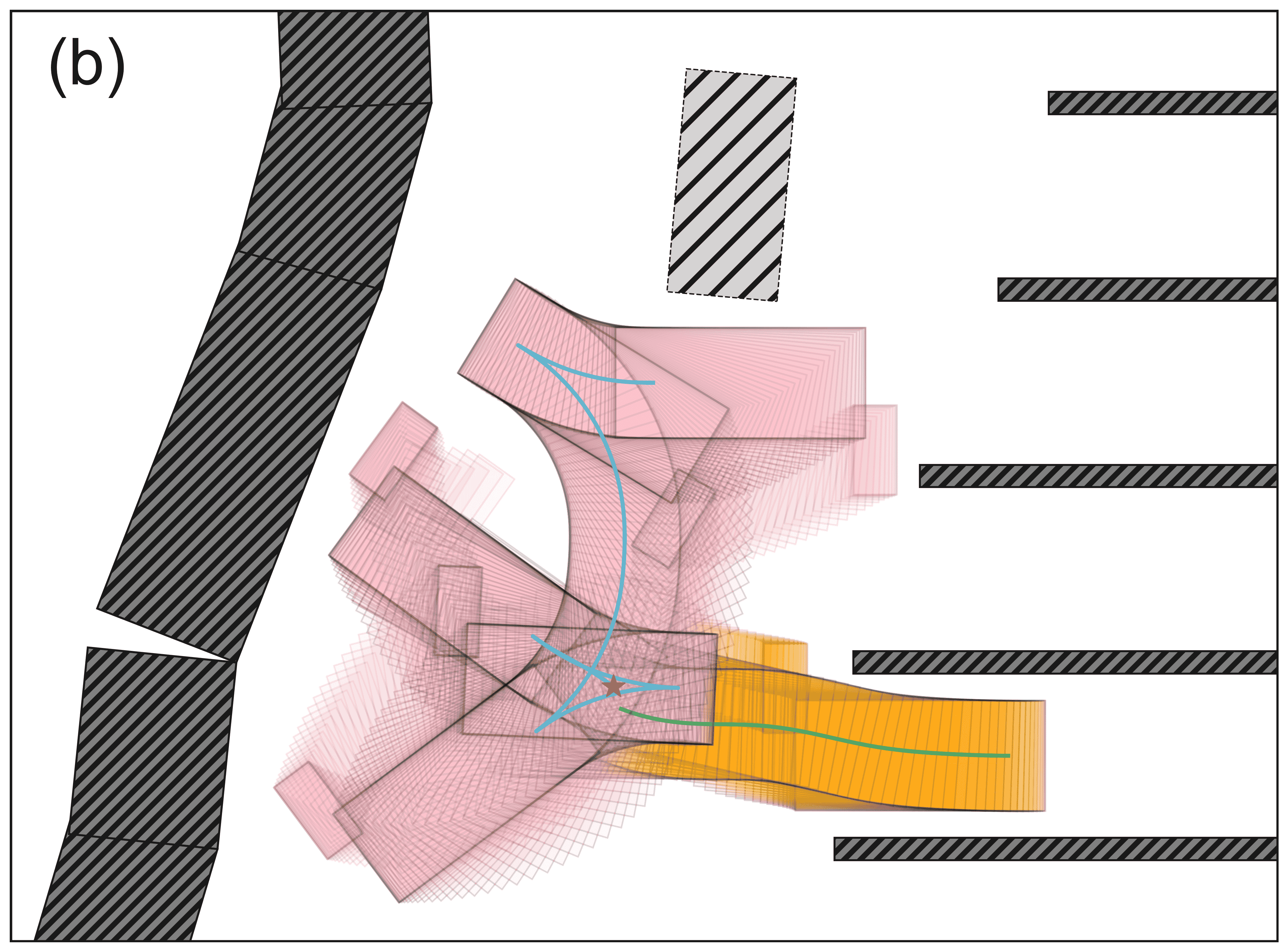}
    \end{subfigure}
    \begin{subfigure}{0.9\linewidth}
        \includegraphics[width=\textwidth]{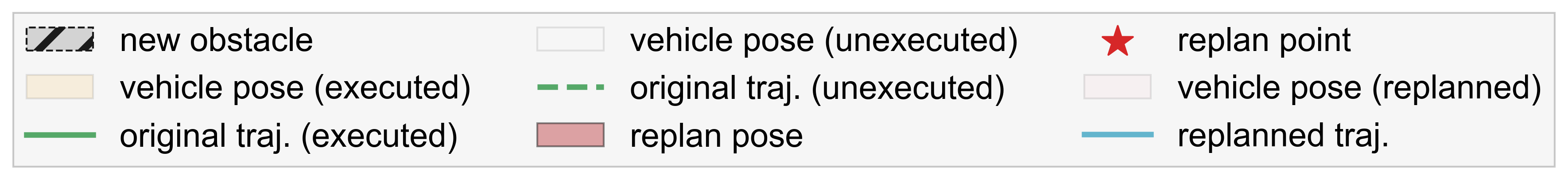}
    \end{subfigure}
    \caption{Illustration of (a) replanning triggered by tracking deviation and newly detected obstacles, and (b) the resulting replanned trajectory.}
\label{fig:replanning_experiment}
\end{figure}

In headland turning, frequent replanning may not be necessary but becomes crucial in situations such as poor trajectory tracking or detecting new obstacles. In Fig.~\ref{fig:replanning_experiment}, we demonstrated that our planner can efficiently generate a new trajectory in such scenarios. Initially, the planner calculated a safe trajectory, but a terrain-induced side slip caused the AAV to deviate to an unintended pose (red star point) during tracking. In addition, a new obstacle (e.g., a tractor), initially excluded from the planning, was detected and posed a collision risk. Replanning was triggered at this point, and assuming the obstacle remained stationary, our planner generated a new trajectory that safely guided the AAV to the target pose. The original trajectory computation took 0.92 s, and the replanned trajectory required only 0.73 s, satisfying real-time efficiency requirements.

\subsection{Results on a Physical Robot}
\label{sec::results_real_robot}

\begin{figure}[t]
\centering
    \includegraphics[width=0.65\linewidth]{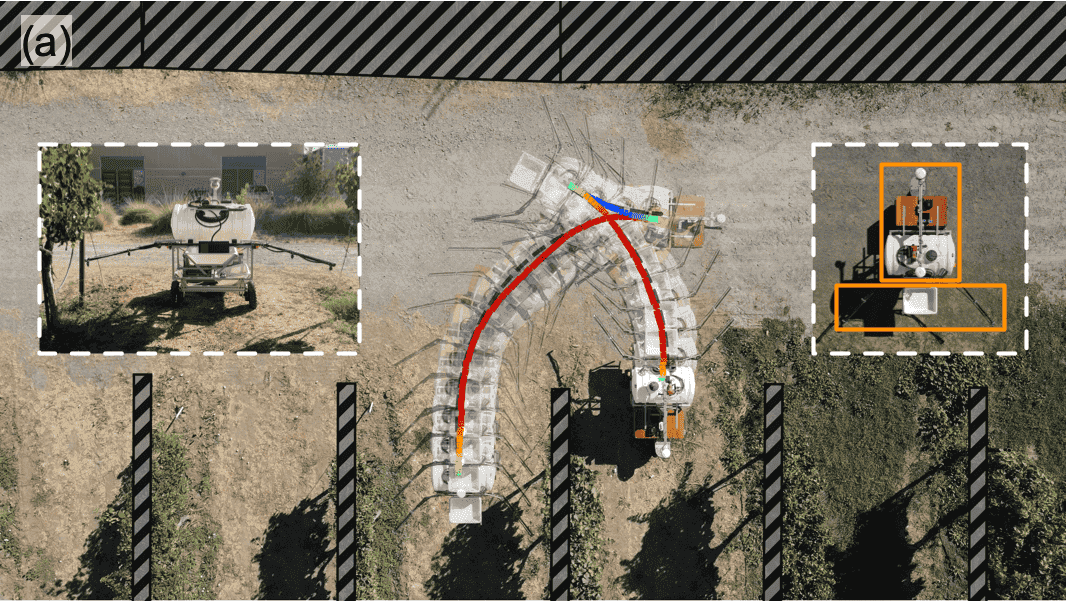}
    \includegraphics[width=0.65\linewidth]{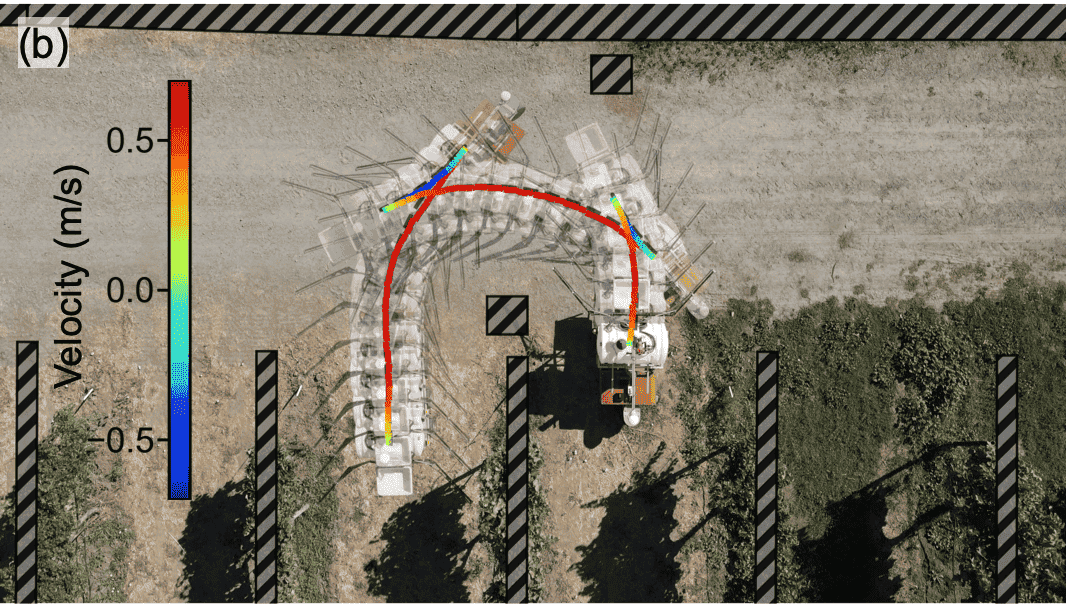}
    \caption{Results on a tricycle robot carrying a spray boom: (a) without obstacles, and (b) with obstacles in the headland space. The geometric representations of the environment and AAV are shown, with planned trajectories as black dashed lines and actual trajectories color-coded by velocity.}
    \label{fig:real_world_experiment}
\end{figure}

To demonstrate the efficacy of the proposed trajectory planner in the real world, we implemented our planning algorithm on a tricycle robot carrying a 3-point spray boom operating in a vineyard. The environment was mapped to include obstacle locations, and a model-predictive control tracker was used to follow the planned trajectories. We tested two scenarios to evaluate performance: (1) without stationary obstacles and (2) with virtual obstacles in the headland space, as shown in Fig.\ref{fig:real_world_experiment}. The obstacle locations, the AAV representation, and the planned and actual trajectories are presented. The total computation time was 0.97s without obstacles and 2.35s with obstacles on an onboard computer with an i5-7300 processor. The results show that our planner can be successfully deployed on a real robot, with the robot accurately following the planned trajectory and safely maneuvering from the initial pose to the target pose in both scenarios. 

In real-world applications, factors like sensor noise, limited fields of view, and dynamic obstacles can impact trajectory planning performance. While we assume RTK GPS for localization, dense foliage may weaken GPS signals. This can be mitigated by integrating visual SLAM algorithms and enhanced by fusing multi-camera or LiDAR inputs for improved coverage and situational awareness. Although we only consider static obstacles in this work, our planner supports real-time replanning, making it adaptable to unmapped or dynamic obstacles. A limitation of our work lies in its reliance on empirically chosen parameters, a common issue in optimization-based approaches. Future work will include a comprehensive sensitivity analysis to evaluate algorithm robustness and explore methods for parameter selection that generalize across scenarios. Additionally, we will extend our approach to accommodate multi-joint or multi-axis vehicles by integrating advanced modeling techniques and adopting modular representations with geometric primitives to support a wider range of agricultural machinery.

\section{CONCLUSION}
\label{sec:conclusion}

This research presents an optimization-based trajectory planner that improves headland turning performance for AAVs, drawing on established practices from both the agricultural domain and robotics. By employing a covering-circle collision detection method and modeling the vehicle-implement geometry with multiple rectangles instead of a single bounding shape, our approach reduces conservatism and enables more efficient maneuvers in constrained headland conditions. Simulation results show an 84.6\% reduction in front-end search time and over 91\% reduction in back-end computation compared to the baseline method. Additionally, our method outperforms a state-of-the-art approach adapted from autonomous driving when applied to conservative agricultural environments. Physical tests in a vineyard further validate the planner’s effectiveness. These findings highlight the practical advantages of enhanced geometric modeling and support broader AAV deployment in complex orchards.



\bibliographystyle{IEEEtran}
\bibliography{IEEEabrv,reference}  

\end{document}